\documentclass[9pt, conference]{IEEEtran} 

\usepackage{cite}
\usepackage{amsmath, amssymb, amsthm}
\usepackage{graphicx}
\usepackage{textcomp}
\usepackage{xcolor}
\usepackage{listings}
\usepackage{booktabs}
\usepackage{makecell}
\usepackage{url}

\makeatletter
\let\MYcaption\@makecaption
\makeatother
\usepackage[font=footnotesize,labelformat=simple]{subcaption}

\makeatletter
\let\@makecaption\MYcaption
\makeatother

\usepackage{svg}
\usepackage{tikz}
\usetikzlibrary{arrows.meta,patterns}

\definecolor{camera}{HTML}{B0BBF0}
\definecolor{cameraPale}{HTML}{E7EBFB}
\definecolor{detector}{HTML}{F8D98C}
\definecolor{detectorPale}{HTML}{FCF2D6}
\definecolor{planner}{HTML}{ADD8AD}
\definecolor{plannerPale}{HTML}{E3F2E3}

\usepackage[ruled,vlined,linesnumbered]{algorithm2e}
\SetAlFnt{\footnotesize}
\SetAlCapFnt{\footnotesize}
\SetAlCapNameFnt{\footnotesize}
\SetKw{Break}{break}

\newtheoremstyle{tightplain}{1pt plus 1pt minus 1pt}{1pt plus 1pt minus 1pt}{\itshape}{}{\bfseries}{.}{ }{}
\newtheoremstyle{tightdef}{1pt plus 1pt minus 1pt}{1pt plus 1pt minus 1pt}{\normalfont}{}{\bfseries}{.}{ }{}
\theoremstyle{tightplain}
\newtheorem{theorem}{Theorem}
\newtheorem{lemma}{Lemma}

\theoremstyle{tightdef}
\newtheorem{example}{Example}

\theoremstyle{plain}
\makeatletter
\renewenvironment{proof}[1][\proofname]{\par
  \pushQED{\qed}\relax
  \normalfont \topsep2\p@\@plus1\p@\@minus1\p@\relax
  \trivlist
  \item[\hskip\labelsep\itshape #1\@addpunct{.}]\ignorespaces
}{\popQED\endtrivlist\@endpefalse}
\makeatother

\usepackage{dblfloatfix}
\usepackage{flafter}

\newcount\phn \newcount\phm
\newcommand\phword{\ifcase\phm lorem\or ipsum\or dolor\or sit\or amet\or sed\or magna\or aliqua\or enim\or minim\fi}
\newcommand{\placeholder}[2]{{\color{black!45}\textbf{[#2]}\phn=0 \loop\ifnum\phn<#1 \phm=\phn \divide\phm by 10 \multiply\phm by -10 \advance\phm by \phn\ \phword\advance\phn by 1 \repeat\par}}

\newcommand{\myparagraph}[1]{\noindent\textbf{#1.}}

\usepackage[hidelinks]{hyperref}

\begin{document}

\title{One Behavior, Out of Order: Deterministic Execution and Parallel Simulation of ROS 2 Applications}

\author{\IEEEauthorblockN{Chengzi Huang, Behnam Khodabandeloo, Duc Anh Nguyen, Chuanchao Gao and Wang Yi}
\IEEEauthorblockA{Department of Information Technology, Uppsala University, Sweden}}

\maketitle

\begin{abstract}
System-level simulation is widely used to validate robotic applications before hardware deployment.
However, Robot Operating System 2 (ROS~2), a popular middleware for such applications, uses asynchronous publish-subscribe communication, which can make application behavior depend on scheduling and the platform.
The behavior can then vary across runs and machines even with the same inputs.
This nondeterminism makes faults hard to reproduce and tests unreliable.
In this paper, we first give ROS~2 applications a lightweight coordination scheme in which every component, including event-driven ones, reads its inputs and writes its outputs at predefined time instants.
By decoupling computation and communication, the scheme makes an application deterministic, that is, it always has one behavior for given inputs. 
Tests are thus repeatable, but for large applications they still take a long time.
To speed up these tests, we then present an out-of-order parallel simulation algorithm.
From the predefined read and write instants, it knows before the run which activations of components depend on which.
It starts each activation as soon as those it depends on have finished.
Finally, we implement the algorithm in a simulator that runs existing ROS~2 applications without any change to them or to ROS~2.
We evaluate the simulator on industrial use cases and synthetic graphs.
On the industrial use cases, it is up to 2.6 times faster than synchronous parallel simulation and needs an order of magnitude less memory for messages than asynchronous parallel simulation.
\end{abstract}

\smallskip
\noindent\textbf{Code}\quad{\urlstyle{tt}\url{https://github.com/ChengziHuang/let_ros}}\par

\section{Introduction}
\label{sec:intro}

\begin{figure}[t]
\centering
\begin{subfigure}[b]{86pt}
\centering
\begin{tikzpicture}[x=1pt, y=1pt, font=\fontsize{6.5}{7.5}\selectfont,
  nd/.style={draw=black!70, rounded corners=2pt, line width=0.4pt},
  cb/.style={draw=black!70, fill=white, rounded corners=1.5pt, line width=0.4pt,
    inner sep=1.2pt, minimum height=8.5pt, minimum width=20pt},
  tm/.style={draw=black!60, fill=white, dashed, rounded corners=1.5pt, line width=0.4pt,
    inner sep=1.2pt, minimum height=8pt, minimum width=20pt},
  st/.style={draw=black!45, fill=black!8, line width=0.4pt, inner sep=1.2pt,
    minimum height=8pt, minimum width=20pt, font=\ttfamily\fontsize{6}{7}\selectfont},
  ar/.style={-{Latex[length=2.6pt,width=2.1pt]}, line width=0.45pt, draw=black!80},
  tp/.style={font=\ttfamily\fontsize{6}{7}\selectfont, inner sep=0.5pt}]
  \draw[nd, fill=cameraPale]   (0,81.5) rectangle (84,96.5);
  \draw[nd, fill=detectorPale] (0,59)   rectangle (84,74);
  \draw[nd, fill=plannerPale]  (0,9.5)  rectangle (84,51.5);
  \node[inner sep=0.5pt] at (14.5,89)   {Camera};
  \node[inner sep=0.5pt] at (14.5,66.5) {Detector};
  \node[inner sep=0.5pt] at (14.5,30.5) {Planner};
  \node[tm] (ct)  at (40,89) {10\,ms};
  \node[cb] (cap) at (69,89) {capture};
  \draw[ar] (ct.east) -- (cap.west);
  \node[cb] (det) at (69,66.5) {detect};
  \node[cb] (sto) at (69,44)   {store};
  \node[st] (lat) at (69,30.5) {latest};
  \node[cb] (pl)  at (69,17)   {plan};
  \node[tm] (pt)  at (40,17)   {20\,ms};
  \draw[ar] (sto.south) -- (lat.north);
  \draw[ar] (lat.south) -- (pl.north);
  \draw[ar] (pt.east) -- (pl.west);
  \draw[ar] (cap.south) -- (det.north);
  \draw[ar] (det.south) -- (sto.north);
  \draw[ar] (pl.south) -- (69,1.5);
  \node[tp, anchor=east] at (67.5,77.75) {/img};
  \node[tp, anchor=east] at (67.5,55.25) {/obj};
  \node[tp, anchor=east] at (67.5,5)     {/cmd};
\end{tikzpicture}\relax\relax
\caption{The application.}
\label{fig:example-app}
\end{subfigure}\hfill
\begin{subfigure}[b]{158pt}
\centering
\begin{tikzpicture}[x=2pt, y=1pt, font=\fontsize{6.5}{7.5}\selectfont,
  ar/.style={-{Latex[length=2.6pt,width=2.1pt]}, line width=0.5pt, draw=black!70}]
  \def\cb#1#2#3#4{\fill[#4] (#1,#3-3.2) rectangle (#2-0.2,#3+3.2);}
  \foreach \t in {10,20,30,40} {\draw[black!12, dotted, line width=0.4pt] (\t,-75.3) -- (\t,11);}
  \draw[-{Latex[length=2.4pt,width=2pt]}, line width=0.4pt] (8,12) -- (46,12);
  \node[anchor=west, inner sep=0.5pt] at (46,12) {ms};
  \foreach \t in {10,20,30,40} {
    \draw[line width=0.4pt] (\t,11) -- (\t,13);
    \node[above, inner sep=1pt] at (\t,13) {\t};}
  \foreach \run/\y in {1/0, 2/-44.8} {
    \node[rotate=90, font=\fontsize{6.5}{7.5}\selectfont\bfseries] at (-8.2,\y-13.5) {Run \run};
    \node[anchor=east, inner sep=1pt] at (8,\y)      {Camera};
    \node[anchor=east, inner sep=1pt] at (8,\y-13.5) {Detector};
    \node[anchor=east, inner sep=1pt] at (8,\y-27)   {Planner};
    \foreach \d in {0,13.5,27} {\draw[black!22, line width=0.3pt] (8,\y-\d) -- (44,\y-\d);}
    \foreach \t in {10,20,30,40} {\cb{\t}{\t+1.2}{\y}{camera}}
    \foreach \t in {20,40} {\cb{\t}{\t+1.2}{\y-27}{planner}}
  }
  \foreach \s/\e/\l in {11/19/1, 21/29/2, 31/39/3} {
    \cb{\s}{\e}{-13.5}{detector} \node at ({(\s+\e)/2},-13.5) {\l};}
  \foreach \s/\e/\l in {11/19/1, 21/33/2, 33/41/3} {
    \cb{\s}{\e}{-58.3}{detector} \node at ({(\s+\e)/2},-58.3) {\l};}
  \draw[ar] (38.9,-16.7) -- (40.2,-23.8);
  \node[anchor=west, inner sep=1pt] at (41.8,-27) {uses image 3};
  \draw[ar] (32.9,-61.5) -- (40.1,-68.6);
  \node[anchor=west, inner sep=1pt] at (41.8,-71.8) {uses image 2};
\end{tikzpicture}\relax\relax
\caption{Two runs with the same inputs.}
\label{fig:example-runs}
\end{subfigure}
\caption{Running example.
Detector's execution time changes the behavior.}
\label{fig:example}
\end{figure}
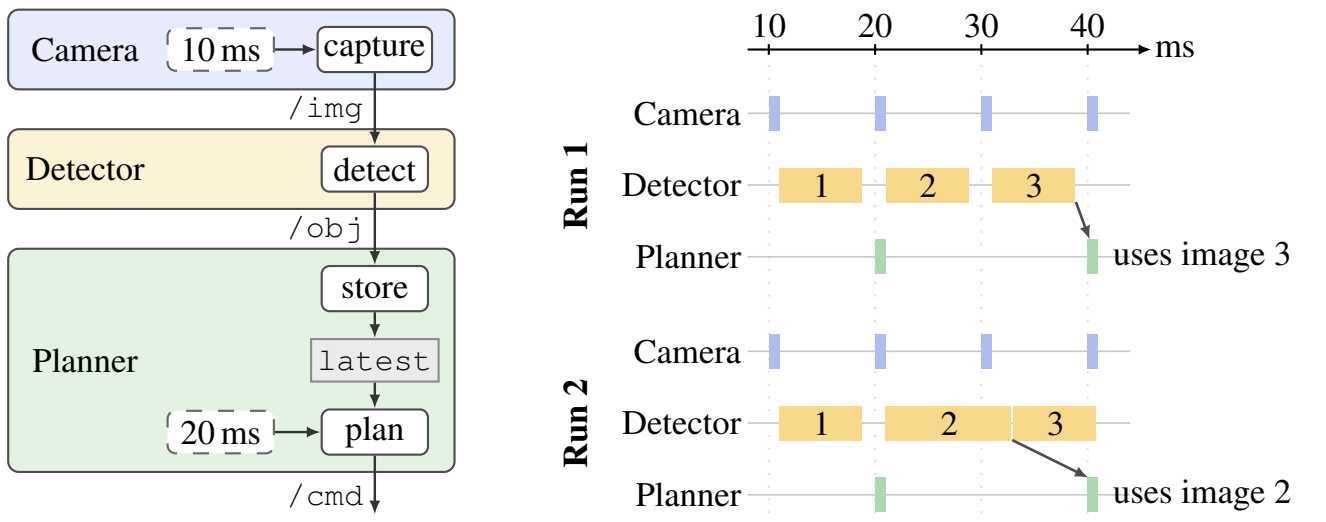

Robot Operating System 2 (ROS~2)~\cite{Macenski2022} is widely used in research and industry to build robotic systems, from mobile robots to automated vehicles.
System-level simulation plays a pivotal role in their development, since developers can validate a whole application across many scenarios without the costly target hardware.
However, the components of a ROS~2 application execute asynchronously and communicate with each other.
The behavior of the application therefore depends on the execution times of its components, the communication delays and the scheduling.
Their variation makes simulation results differ across runs, even when the application code and inputs remain unchanged~\cite{Lee2006, Lee2021}.
In the example of Fig.~\ref{fig:example}, Planner computes a command every 20\,ms from the latest objects that Detector found in the camera images.
In run~2, a variation in the execution time of Detector leads Planner to compute the command at 40\,ms from image~2 instead of image~3.
If the command computed from image~2 causes a fault, the fault may not appear when the test is repeated, which makes it hard to reproduce.
Developers may therefore need to rerun the same test many times to reproduce and diagnose the fault. 
This additional debugging effort slows development.

AUTOSAR solves this problem for automotive software with a rule by which every component reads its inputs and writes its outputs at predefined time instants, no matter when the computations run in between~\cite{AUTOSARTimex}.
This separates computation from communication~\cite{Keutzer2000, Hamann2017}, and the behavior of an application becomes independent of the scheduling and the platform.
Following the same principle, our first step gives ROS~2 a lightweight coordination scheme that fixes when each component reads and writes, including event-driven components.
Whenever a time-triggered or event-driven component is activated, it reads its inputs at that instant and writes its outputs after a fixed computation delay, even if they are empty.
Every output arrives after a fixed transport delay.
Since time-triggered components run periodically and the delays are fixed, all reads and writes happen at predefined time instants, as in AUTOSAR.
The application therefore becomes \textit{deterministic} and exhibits \emph{one behavior}.
Tests become repeatable, and application behavior can be validated independently of the target hardware.

However, system-level simulation still becomes time-consuming as applications and scenarios grow, slowing design iteration~\cite{Antinyan2020}.
Even with one behavior, a recent simulator for ROS~2 applications runs all activations on a single thread, which its authors note increases the run time of complex simulations~\cite{Sagmeister2026}.
To further reduce the validation time, we therefore aim to exploit parallelism within a single simulation while preserving the application's deterministic behavior.
Yet general parallel discrete-event simulation cannot assume that event dependencies are known in advance, so it must preserve causality during the run, either synchronously or asynchronously.
Synchronous parallel simulation stops all workers at a global barrier after each time instant or window~\cite{Fujimoto2000}, forcing fast workers to wait for the slowest one.
Asynchronous parallel simulation avoids global barriers, either through optimistic execution, which needs state saving and rollback~\cite{Jefferson1985}, or through conservative execution with null messages~\cite{ChandyMisra1979}.
When fast producers run far ahead of their consumers, it can accumulate large buffers of unconsumed messages.

Out-of-order simulation needs none of these mechanisms if the dependencies between events are known.
Existing out-of-order simulators derive them from a static analysis of the model or track them during the run~\cite{ChenDomer2014,Jensen2025,Xie2025}.
Our key observation is that the coordination scheme already makes these dependencies known in advance, without any such analysis or tracking.
The predefined read and write instants determine which outputs each activation reads.
An activation therefore need not wait for all earlier activations.
Its computation depends only on the activations whose outputs it reads and on the previous activation of the same component, whose updated state it needs.
This allows activations of different components to run \textit{out of order} and in \textit{parallel} while preserving the one behavior of the application.
Building on this observation, we present an out-of-order parallel simulation algorithm for ROS~2 applications.
It computes the dependencies on demand from the application graph and the timing parameters, without constructing the full dependency graph.
Because a producer may run ahead of its consumers, several outputs may need to be retained at the same time.
To bound output buffering, the algorithm uses cyclic storage and allows a slot to be reused only after all readers of its previous contents have finished.

Together, the coordination scheme and the simulator make simulation-based validation of ROS~2 applications both repeatable and fast, which shortens each design iteration.
The main contributions of this paper are as follows.
\begin{itemize}
\item A lightweight coordination scheme that makes ROS~2 applications deterministic, so that their behavior does not depend on the scheduling or the platform (Section~\ref{sec:model}).
\item An efficient out-of-order parallel simulation algorithm that still computes the one deterministic behavior of an application (Section~\ref{sec:algorithm}).
We show how to integrate it into ROS~2 so that it simulates existing applications without changing their source code or the libraries of ROS~2 (Section~\ref{sec:layer}).
\item We evaluate our simulator on industrial use cases and synthetic graphs.
On 256 cores, it is up to $228\times$ faster than sequential simulation, which is 91\,\% of the theoretical speedup bound.
Compared with synchronous parallel simulation, it is up to $2.6\times$ faster on the industrial use cases and about $2\times$ faster on the synthetic graphs.
Its message buffers stay much smaller than those of asynchronous parallel simulation (Section~\ref{sec:evaluation}).
\end{itemize}

\section{System Model and Execution Semantics}
\label{sec:model}

We model a ROS~2 application as a graph and define our coordination scheme for deterministic execution.

\subsection{System Model}
\label{subsec:graph}

From a logical perspective, a ROS~2 application is a set of \emph{nodes} that communicate using the publish-subscribe paradigm~\cite{Macenski2022}.
Nodes publish messages on named \emph{topics} and subscribe to topics to receive messages.
A node maintains state in member variables and may have \emph{timers} and \emph{subscriptions}, each associated with a \emph{callback}, a function invoked in response to an event.
Timer callbacks are \emph{time-triggered}, activated periodically, whereas subscription callbacks are \emph{event-driven}, activated by incoming messages.
Callbacks may update node state and publish messages that activate other callbacks, forming a network of topics and callbacks.\footnote{ROS~2 runs ready callbacks on one or more threads.
Note that, by default, it never runs two callbacks of the same node at the same time~\cite{ros2cbg}.}
In Fig.~\ref{fig:example-app}, \texttt{capture} and \texttt{plan} are time-triggered, and \texttt{detect} and \texttt{store} are event-driven.

We model an application as a finite directed graph $G = (V, E)$.
Each vertex $z \in V$ is a callback, and each edge in $E \subseteq V \times V$ is a \emph{communication channel}.
An event-driven callback $u$ has one \emph{input topic}, the topic of its subscription.
Every callback $z$ has a fixed finite \emph{output set} $O_z$, which contains the topics on which $z$ may publish.
A pair $e = (z, u)$ is in $E$ if and only if $u$ is event-driven and its input topic is in $O_z$.

The callbacks of one node form a \emph{task}.
Each task $\tau_i$ has a state $s_i$ with initial value $s_{i,0}$, a list $X_i$ of time-triggered callbacks and a list $U_i$ of event-driven callbacks.
The state models the member variables of the node, and only the callbacks of this task can read and update it.
Each time-triggered callback $x \in X_i$ has a period $T_x \in \mathbb{Z}_{>0}$ and an offset $\phi_x \in \mathbb{Z}_{\ge 0}$, and it is activated at the instants $\phi_x + m T_x$ for $m\in\mathbb{Z}_{\ge 0}$.
A callback $z$ of $\tau_i$ computes a function $f_z$.
Its inputs are the state $s_i$, the current time and, if $z$ is event-driven, the incoming message.
Its outputs are the new value of $s_i$ and the messages that $z$ publishes on the topics in $O_z$.
We assume that $f_z$ is \emph{deterministic}, that is, its outputs depend only on its inputs.

\begin{figure*}[t]
\centering
\begin{subfigure}[b]{112pt}
\centering
\begin{tikzpicture}[x=1pt, y=1pt, font=\fontsize{7}{8}\selectfont,
  tk/.style={draw=black!70, rounded corners=2pt, line width=0.4pt},
  ed/.style={circle, draw=black!75, fill=white, line width=0.4pt, inner sep=0pt,
    minimum size=11pt},
  tt/.style={ed, double, double distance=0.8pt, minimum size=9.4pt},
  st/.style={draw=black!45, fill=black!8, line width=0.4pt, inner sep=0pt,
    minimum height=7.2pt, minimum width=11pt},
  ch/.style={-{Latex[length=2.6pt,width=2.1pt]}, line width=0.45pt, draw=black!80},
  sh/.style={line width=0.4pt, draw=black!45},
  lb/.style={anchor=west, inner sep=0pt},
  pr/.style={lb}]
  \useasboundingbox (-0.2,0) rectangle (108.2,81.92);
  \draw[tk, fill=cameraPale]   (0,67.52) rectangle (108,81.72);
  \draw[tk, fill=detectorPale] (0,45.17) rectangle (108,59.37);
  \draw[tk, fill=plannerPale]  (0,0.22)  rectangle (108,37.02);
  \node[lb] at (2.5,74.62) {$\tau_1$};  \node[pr] at (11.5,74.62) {$D_1\mkern1.5mu{=}\mkern1.5mu 10$};
  \node[lb] at (2.5,52.27) {$\tau_2$};  \node[pr] at (11.5,52.27) {$D_2\mkern1.5mu{=}\mkern1.5mu 25$};
  \node[lb] at (2.5,29.92) {$\tau_3$};  \node[pr] at (11.5,29.92) {$D_3\mkern1.5mu{=}\mkern1.5mu 20$};
  \node[tt] (x1) at (46.5,74.62) {$x_1$};
  \node[ed] (u2) at (46.5,52.27) {$u_2$};
  \node[ed] (u3) at (46.5,29.92) {$u_3$};
  \node[st] (s3) at (46.5,18.62) {$s_3$};
  \node[tt] (x3) at (46.5,7.32)  {$x_3$};
  \node[pr] at (55.5,74.62) {$T_{x_1}\mkern1.5mu{=}\mkern1.5mu \phi_{x_1}\mkern1.5mu{=}\mkern1.5mu 10$};
  \node[pr] at (55.5,7.32)  {$T_{x_3}\mkern1.5mu{=}\mkern1.5mu \phi_{x_3}\mkern1.5mu{=}\mkern1.5mu 20$};
  \draw[sh] (u3.south) -- (s3.north);
  \draw[sh] (s3.south) -- (x3.north);
  \draw[ch] (x1.south) -- (u2.north);
  \draw[ch] (u2.south) -- (u3.north);
  \node[lb] at (51,63.45) {$e_1$};  \node[pr] at (62,63.45) {$\delta_{e_1}\mkern1.5mu{=}\mkern1.5mu 0$};
  \node[lb] at (51,41.1)  {$e_2$};  \node[pr] at (62,41.1)  {$\delta_{e_2}\mkern1.5mu{=}\mkern1.5mu 4$};
\end{tikzpicture}\relax\relax
\caption{Graph model.}
\label{fig:example-graph}
\end{subfigure}\hfill
\begin{subfigure}[b]{186pt}
\centering
\begin{tikzpicture}[x=2pt, y=1pt, font=\fontsize{6.5}{7.5}\selectfont,
  jl/.style={font=\fontsize{7}{8}\selectfont, inner sep=0pt},
  rc/.style={-{Latex[length=2.6pt,width=2.1pt]}, line width=0.5pt, draw=black!70},
  er/.style={rc, draw=black!55, dash pattern=on 1.6pt off 1pt}]
  \def\win#1#2#3#4{\fill[#4] (#1+0.25,#3-3.5) rectangle (#2-0.25,#3+3.5);}
  \def\run#1#2#3#4{\fill[#4] (#1,#3-3.5) rectangle (#2,#3+3.5);}
  \foreach \t in {10,20,...,70} {\draw[black!12, dotted, line width=0.4pt] (\t,-64) -- (\t,9);}
  \draw[-{Latex[length=2.4pt,width=2pt]}, line width=0.4pt] (8,10) -- (76,10);
  \node[anchor=west, inner sep=0.5pt] at (76,10) {ms};
  \foreach \t in {10,20,...,70} {
    \draw[line width=0.4pt] (\t,9) -- (\t,11);
    \node[above, inner sep=1pt] at (\t,11) {\t};}
  \node[anchor=east, inner sep=1pt] at (8,0)     {Camera {\fontsize{7}{8}\selectfont$\tau_1$}};
  \node[anchor=east, inner sep=1pt] at (8,-21.5) {Detector {\fontsize{7}{8}\selectfont$\tau_2$}};
  \node[anchor=east, inner sep=1pt] at (8,-51.5) {Planner {\fontsize{7}{8}\selectfont$\tau_3$}};
  \foreach \y in {0,-13,-21.5,-30,-43,-51.5,-60} {\draw[black!22, line width=0.3pt] (8,\y) -- (75,\y);}
  \foreach \y in {0,-21.5,-51.5} {\node[anchor=west, inner sep=0.5pt] at (75.6,\y) {$\cdots$};}
  \foreach \k/\t/\s/\e/\l in {1/10/16.9/19.2/13.1, 2/20/26.9/28.3/23.1, 3/30/36.9/39.6/33.1,
                              4/40/47.3/49.2/43.1, 5/50/56.9/57.9/53.1, 6/60/66.9/69.3/63.1} {
    \win{\t}{\t+10}{0}{cameraPale} \run{\s}{\e}{0}{camera}
    \node[jl] at (\l,0) {$J_{1,\k}$};}
  \win{20}{45}{-13}{detectorPale}      \run{22.5}{29.2}{-13}{detector} \node[jl] at (25.8,-13) {$J_{2,1}$};
  \win{50}{75.25}{-13}{detectorPale}   \run{61}{69}{-13}{detector}   \node[jl] at (65,-13) {$J_{2,4}$};
  \win{30}{55}{-21.5}{detectorPale}    \run{32.5}{37.5}{-21.5}{detector} \run{41}{44.5}{-21.5}{detector}
                                    \node[jl] at (35,-21.5) {$J_{2,2}$};
  \win{60}{75.25}{-21.5}{detectorPale} \run{70}{75}{-21.5}{detector} \node[jl] at (65,-21.5) {$J_{2,5}$};
  \win{40}{65}{-30}{detectorPale}      \run{48.5}{54.5}{-30}{detector} \node[jl] at (51.5,-30) {$J_{2,3}$};
  \win{49}{69}{-43}{plannerPale}       \run{52}{53.5}{-43}{planner}   \node[jl] at (63,-43) {$J_{3,3}$};
  \draw[planner!60!black, dash pattern=on 1.4pt off 0.9pt, line width=0.45pt]
    (75,-46.5) -- (69.25,-46.5) -- (69.25,-39.5) -- (75,-39.5);
  \node[jl, text=black!60] at (72.4,-43) {\scalebox{0.86}{$J_{3,6}$}};
  \win{59}{75.25}{-51.5}{plannerPale}  \run{62}{64.5}{-51.5}{planner} \node[jl] at (69,-51.5) {$J_{3,4}$};
  \win{20}{40}{-60}{plannerPale}       \run{30.5}{35.5}{-60}{planner} \node[jl] at (25,-60) {$J_{3,1}$};
  \win{40}{60}{-60}{plannerPale}       \run{43.5}{46.5}{-60}{planner} \node[jl] at (52.5,-60) {$J_{3,2}$};
  \win{60}{75.25}{-60}{plannerPale}    \run{67}{73}{-60}{planner}     \node[jl] at (63.2,-60) {$J_{3,5}$};
  \foreach \t/\y in {20/-9.5, 30/-18, 40/-26.5, 50/-9.5, 60/-18} {\draw[rc] (\t,-3.5) -- (\t,\y);}
  \draw[rc] (45,-16.5) -- (49,-39.5);
  \draw[rc] (55,-25) -- (59,-48);
  \draw[er] (65,-33.5) -- (69,-39.5);
\end{tikzpicture}\relax\relax
\caption{Execution timeline.}
\label{fig:example-rule}
\end{subfigure}\hfill
\begin{subfigure}[b]{166pt}
\centering
\begin{tikzpicture}[x=2.1pt, y=1pt, font=\fontsize{7}{8}\selectfont,
  disc/.style={circle, draw=none, inner sep=0pt, minimum size=15pt},
  lab/.style={inner sep=0pt, font=\fontsize{7}{8}\selectfont},
  r1/.style={-{Latex[length=2pt,width=1.6pt]}, line width=0.35pt, draw=black!40},
  r2/.style={-{Latex[length=2.4pt,width=1.9pt]}, line width=0.45pt, draw=black!75},
  r2e/.style={r2, draw=black!55, dash pattern=on 1.6pt off 1pt},
  r3/.style={-{Latex[length=2.4pt,width=1.9pt]}, line width=0.45pt, draw=red!75!black,
    dash pattern=on 1.5pt off 0.9pt}]
  \useasboundingbox (1.2,-74.42) rectangle (78,7.5);
  \foreach \y/\s in {0/x_1, -21.3/u_2, -45.6/u_3, -66.92/x_3} {
    \node[anchor=east, inner sep=1pt] at (5.4,\y) {$\s$};}
  \foreach \k in {1,...,6} {\node[disc, fill=cameraPale] (c\k) at (10*\k,0) {}; \node[lab] at (c\k) {$J_{1,\k}$};}
  \foreach \k in {1,...,5} {\node[disc, fill=detectorPale] (d\k) at (10*\k+10,-21.3) {}; \node[lab] at (d\k) {$J_{2,\k}$};}
  \foreach \k/\t/\y in {1/20/-66.92, 2/40/-66.92, 5/60/-66.92, 3/49/-45.6, 4/59/-45.6} {
    \node[disc, fill=plannerPale] (p\k) at (\t,\y) {}; \node[lab] at (p\k) {$J_{3,\k}$};}
  \node[disc, draw=planner!60!black, dash pattern=on 1.4pt off 0.9pt, line width=0.45pt,
    minimum size=14.55pt] (p6) at (69,-45.6) {};
  \node[lab, text=black!60] at (p6) {$J_{3,6}$};
  \foreach \x/\y in {67/0, 67/-21.3, 76/-45.6, 67/-66.92} {\node[inner sep=0pt] at (\x,\y) {$\cdots$};}
  \foreach \k [evaluate=\k as \n using int(\k+1)] in {1,...,5} {\draw[r1] (c\k) -- (c\n);}
  \foreach \k [evaluate=\k as \n using int(\k+1)] in {1,...,4} {\draw[r1] (d\k) -- (d\n);}
  \draw[r1] (p1) -- (p2);
  \draw[r1] (p2) -- (p3);
  \draw[r1] (p3) -- (p4);
  \draw[r1] (p4) -- (p5);
  \draw[r1] (p5) -- (p6);
  \foreach \k in {1,...,5} {\draw[r2] (c\k) -- (d\k);}
  \draw[r2]  (d1.south east) -- (p3.north west);
  \draw[r2]  (d2.south east) -- (p4.north west);
  \draw[r2e] (d3.south east) -- (p6.north west);
\end{tikzpicture}\relax\relax
\caption{Job dependencies.}
\label{fig:example-deps}
\end{subfigure}
\caption{Graph model, execution timeline and job dependencies of the running example.
(a) Circles are callbacks, with a double line if they are time-triggered.
Arrows are channels, boxes are tasks, and the grey box is the state of $\tau_3$.
(b) A light bar spans from the release instant to the write instant of a job, and an arrow goes from the write of a record to its arrival.
Darker boxes show when the jobs run in one run.
The dashed record is empty and releases the skipped job $J_{3,6}$.
(c) Grey arrows are state dependencies and black arrows are record dependencies.}
\label{fig:sem}
\end{figure*}
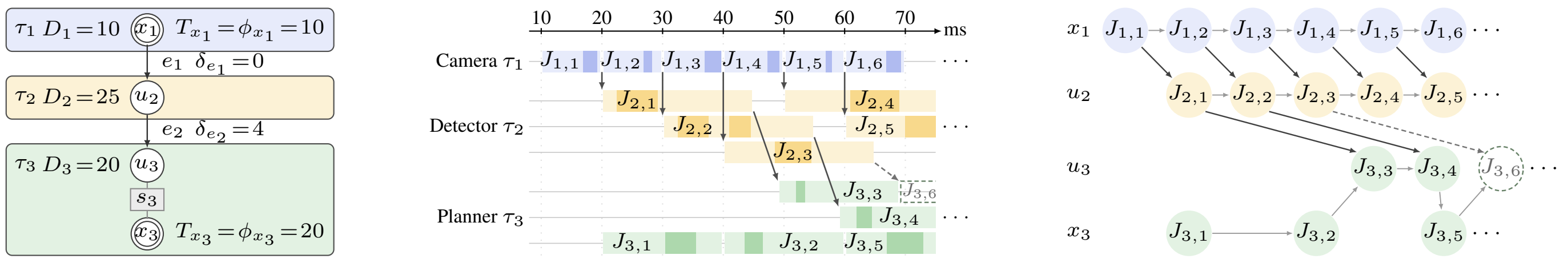

\begin{example}
\label{ex:graph}
Fig.~\ref{fig:example-graph} shows the graph of the application of Fig.~\ref{fig:example-app}, with $x_1$, $u_2$, $u_3$ and $x_3$ for \texttt{capture}, \texttt{detect}, \texttt{store} and \texttt{plan}.
The output sets are $O_{x_1} = \{\texttt{/img}\}$, $O_{u_2} = \{\texttt{/obj}\}$, $O_{x_3} = \{\texttt{/cmd}\}$ and $O_{u_3} = \emptyset$.
The input topics of $u_2$ and $u_3$ are \texttt{/img} and \texttt{/obj}, so $E$ has only $e_1 = (x_1, u_2)$ and $e_2 = (u_2, u_3)$.
The state $s_3$ holds \texttt{latest}, which $u_3$ writes and $x_3$ reads.
\hfill$\diamond$
\end{example}

\subsection{Execution Semantics}
\label{subsec:semantics}

To make the behavior of an application independent of the scheduling and the platform, we separate computation from communication by fixing the instants at which callbacks read their input messages and write their output messages.
For this, every task $\tau_i$ has a \emph{computation delay} $D_i \in \mathbb{Z}_{>0}$ and every channel $e$ has a \emph{transport delay} $\delta_e \in \mathbb{Z}_{\ge 0}$, both predefined and fixed.
Channels from callbacks of the same task to the same callback have the same transport delay.

Every callback releases a sequence of \emph{jobs}.
A job of a callback of $\tau_i$, which we also call a job of $\tau_i$, has a \emph{release instant} and a \emph{write instant}, which is $D_i$ later.
It reads its input messages at its release instant and writes its output messages at its write instant, no matter when it runs in between.
It writes them in \emph{records}, one on every topic in the output set of its callback.
A record holds the messages that the job publishes on its topic, in the order in which it publishes them.
It is \emph{empty} if the job publishes nothing on the topic, for example because its callback publishes only for some inputs.

A time-triggered callback releases a job at every instant at which it is activated.
A record that a job of $z$ writes on the input topic of $u$ enters the channel $e = (z, u)$ when it is written and arrives at $u$ at its \emph{arrival instant}, which is $\delta_e$ later.
At this instant it releases a job of $u$, and only this job \emph{consumes} it, that is, removes it from the channel and reads its messages.
Since $\delta_e$ is fixed, every channel is a first-in first-out queue of records.
An empty record releases a job like any other record, so whether a job is released does not depend on what the callbacks compute.

In ROS~2, the executor decides in which order the ready callbacks of a node run, and the default executor runs timer callbacks first~\cite{ros2executors,rclcppJazzy}.
We fix this order by a rule of our own.
All jobs of $\tau_i$ form one sequence in the order of their release instants, and $J_{i,k}$ denotes the $k$-th of them.
For jobs with the same release instant, jobs of time-triggered callbacks come first, in the order of $X_i$, and jobs of event-driven callbacks follow, in the order of $U_i$.
Jobs of the same event-driven callback are ordered by the jobs that wrote their records, first by the index of the task and then by the index of the job.

The jobs of $\tau_i$ run one after the other in this sequence.
Each job starts from the state that the previous job left, even if it is released before the previous job writes, and the first job starts from $s_{i,0}$.
Whenever a job reads the current time, it gets its release instant.
A job of a time-triggered callback $x$ calls $f_x$ once.
A job of an event-driven callback $u$ calls $f_u$ once for each message of the record that it consumes, in the order of the messages.
If this record is empty, the job is \emph{skipped}, that is, it calls nothing, keeps the state and writes only empty records.
So an empty record comes from a job that publishes nothing on a topic or from a skipped job.

\begin{example}
\label{ex:timeline}
Fig.~\ref{fig:example-rule} shows the jobs of the running example with the parameters of Fig.~\ref{fig:example-graph}.
Job $J_{2,2}$, released at $30$\,ms, runs in two pieces but still writes its record on \texttt{/obj} at $30 + D_2 = 55$\,ms.
This record arrives at $55 + \delta_{e_2} = 59$\,ms and releases $J_{3,4}$, which stores the detection in $s_3$.
The next job $J_{3,5}$ of $\tau_3$ is released at $60$\,ms, so it computes its command from this detection in every run.
Job $J_{2,3}$ finds no object, so its record is empty and the job $J_{3,6}$ that it releases is skipped.
\hfill$\diamond$
\end{example}

\subsection{Determinism}
\label{subsec:properties}

\begin{theorem}
\label{thm:det}
The jobs, their release and write instants, their order in each task and which record each job reads depend only on the application graph model, the tasks, the periods, the offsets and the delays.
The states and the records depend in addition only on the initial states and the functions $f_z$.
So the behavior of the application, that is, its non-empty records with their write instants, depends neither on the scheduling nor on the platform.
\end{theorem}
\begin{proof}
Jobs are released only at predefined instants, so we use induction on the release instant $t$.
Since $D_i > 0$, a job released at $t$ is released by a time-triggered callback or by a record of a job released before $t$, and this record is written whether it is empty or not.
So the jobs released at $t$, their order and the records that they read do not depend on what the callbacks compute.
We take the jobs of each task in the order of its sequence.
Their inputs then depend only on the initial states and on the outputs of jobs that come before them, and so do their outputs, since every $f_z$ is deterministic.
\end{proof}

\subsection{Discussion}
\label{subsec:discussion}

The main benefit of empty records is that the jobs and their dependencies are known before the run, as Theorem~\ref{thm:det} states.
Without them, a callback that publishes only for some inputs would release jobs of other callbacks only for these inputs, so the jobs would depend on the computed values.
With them, every job depends only on the previous job of its task and on the job that wrote the record that it consumes, and both come before it.
So the dependencies form a \textit{directed acyclic graph}, even if the application graph model has cycles.
Fig.~\ref{fig:example-deps} shows this graph for the running example.
Section~\ref{sec:algorithm} shows how to run jobs out of order and in parallel by computing their dependencies from the parameters, without building the graph.

\section{Out-of-Order Parallel Simulation}
\label{sec:algorithm}

As discussed in Section~\ref{subsec:discussion}, the dependencies between jobs form a directed acyclic graph.
A simulation that runs the jobs in a topological order of this graph is \emph{correct}, that is, every job writes the same records as in the semantics of Section~\ref{subsec:semantics}.
So a simulator can run jobs \textit{out of the order} of their release instants and in parallel, as long as every job starts only after the jobs that it depends on have finished.

\begin{example}
\label{ex:parallel}
In Fig.~\ref{fig:example-deps}, $J_{3,5}$ depends, directly or through other jobs, only on $J_{1,1}$, $J_{1,2}$, $J_{2,1}$, $J_{2,2}$ and $J_{3,1}, \dots, J_{3,4}$.
So the simulator can run $J_{3,5}$ in parallel with $J_{1,3}$, or even finish it before $J_{1,3}$ starts, although $J_{3,5}$ is released at $60$\,ms and $J_{1,3}$ at $30$\,ms.
\hfill$\diamond$
\end{example}

\subsection{Activation Streams}
\label{subsec:streams}

To execute jobs in a topological order, the simulator needs to identify their dependencies.
We can compute these dependencies on demand using only the application graph model parameters, without constructing or storing the full job dependency graph.
To this end, we group the jobs into periodic sequences, in which the record and state dependencies of a job (Fig.~\ref{fig:example-deps}) follow from its position.

A time-triggered callback releases a job in every period.
Each of these jobs writes a record on every topic in its output set, empty or not, and after fixed delays each record releases one job of every callback that receives it.
So these jobs are periodic as well, with the same period and a later offset, and so are the jobs that their records release.
We call such a periodic sequence of jobs of one callback an \emph{activation stream}.
A stream $\sigma$ has a period $T_\sigma$ and an offset $\phi_\sigma$, and its $m$-th job $J(\sigma, m)$, with $m\in\mathbb{Z}_{\ge 0}$, is released at $r_\sigma(m) = \phi_\sigma + m T_\sigma$.
Every time-triggered callback $x$ has a \emph{root stream} with period $T_x$ and offset $\phi_x$.
For each channel $e=(z,u)$, every stream $\sigma$ of callback $z$ in task $\tau_j$ has a \emph{child stream} $\nu$ of callback $u$.
The child stream has period $T_\nu=T_\sigma$ and offset $\phi_\nu=\phi_\sigma+D_j+\delta_e$.
We call $\sigma$ the \emph{parent} of $\nu$.
Job $J(\nu,m)$ consumes the record written by $J(\sigma,m)$, so $J(\sigma,m)$ is its record dependency.

Assume that we simulate the jobs released in the interval $[0, H)$, for a given $H \in \mathbb{Z}_{>0}$.
Since $D_j > 0$, a child stream has a later offset than its parent, so only finitely many streams have an offset before $H$.
Such a stream $\sigma$ has $M_\sigma = \lceil (H - \phi_\sigma) / T_\sigma \rceil$ jobs in $[0, H)$.

The state dependency of a job is the previous job in the sequence of its task.
This sequence merges the jobs of all streams of the task and orders them by release instant, which $r_\sigma(m)$ gives.
Jobs of a task with the same release instant belong to different streams.
To order them, we give every stream $\sigma$ of $\tau_i$ with an offset before $H$ a \emph{rank} $q_\sigma$, its position in the following order, which follows the rule of Section~\ref{subsec:semantics}.
Root streams come first, in the order of $X_i$.
Child streams follow, ordered by the position of their callback in $U_i$, then by the index of the task of their parent, and then by the rank of their parent.
This is not circular, since a parent has an earlier offset than its child.
The \emph{key} of $J(\sigma, m)$ is $\kappa_\sigma(m) = (r_\sigma(m), q_\sigma)$, compared first by release instant and then by rank.
By the following lemma, the keys give the sequence of the task, so the state dependency of a job is the job of the same task with the next smaller key.

\begin{lemma}
\label{lem:streams}
Every job released in $[0, H)$ is the job $J(\sigma, m)$ of one stream $\sigma$ and one $m < M_\sigma$.
In the order of their keys, the jobs of $\tau_i$ released in $[0, H)$ are the first jobs $J_{i,1}, J_{i,2}, \dots$ of its sequence.
\end{lemma}
\begin{proof}
By induction on the release instant, every job is in one stream, since it is released by a time-triggered callback or by the record of one earlier job.
The ranks follow the rule of Section~\ref{subsec:semantics}, with the parents of two streams in place of the jobs that wrote the records.
When these jobs are in the same task, their channels have the same transport delay, so they have the same release instant, and by induction their order is the order of the ranks of the parents.
\end{proof}

\begin{example}
\label{ex:streams}
With $H = 70$\,ms, the running example has four streams.
The root stream $\sigma_1$ of $x_1$ has period and offset $10$\,ms, and its child $\sigma_2$ is a stream of $u_2$ with offset $20$\,ms.
The child $\sigma_3$ of $\sigma_2$ is a stream of $u_3$ with offset $20 + D_2 + \delta_{e_2} = 49$\,ms.
The root stream $\sigma_4$ of $x_3$ has period and offset $20$\,ms.
In $\tau_3$, $\sigma_4$ has rank $1$ and $\sigma_3$ has rank $2$, so the keys $(20,1)$, $(40,1)$, $(49,2)$, $(59,2)$, $(60,1)$, $(69,2)$ give $J_{3,1}, \dots, J_{3,6}$.
\hfill$\diamond$
\end{example}

\subsection{Bounded Record Storage}
\label{subsec:rings}

\begin{algorithm}[t]
\caption{Out-of-order parallel simulation.}
\label{alg:ooo}
\DontPrintSemicolon
\SetArgSty{textnormal}
\SetKwProg{Proc}{procedure}{}{}
\Proc{\textsc{TryEnqueue}$(\tau_i)$}{
  $k \leftarrow c_i$\;
  \lIf{$a_i \neq k$ \textbf{or} $\tau_i$ has no next job}{\Return}
  $J(\sigma, m) \leftarrow$ the next job of $\tau_i$\;
  \lIf{$\sigma$ has a parent $\pi$ in $\tau_j$ \textbf{and} $\kappa_\pi(m) > \lambda_j$}{\Return}
  \lIf{$m \ge K_\sigma$ \textbf{and} $\sigma$ has a child $\nu$ in $\tau_l$ with $\kappa_\nu(m - K_\sigma) > \lambda_l$}{\Return}
  \lIf{$\mathrm{CAS}(a_i, k, k + 1)$}{push $J(\sigma, m)$ onto the local deque}
}
\Proc{\textsc{Worker}$()$}{
  \While{some task has a next job}{
    $J(\sigma, m) \leftarrow$ a job from the local deque or a stolen job\;
    \lIf{there is no such job}{\textbf{continue}}
    run $J(\sigma, m)$ of $\tau_i$ and write its records into slot $m \bmod K_\sigma$\;
    replace $J(\sigma, m)$ in the heap of $\tau_i$ by $J(\sigma, m + 1)$, or remove it if $m + 1 = M_\sigma$\;
    set the next job of $\tau_i$ to the first job in its heap, if any\;
    $\lambda_i \leftarrow \kappa_\sigma(m)$, then $c_i \leftarrow c_i + 1$\;
    \lFor{$\tau_j = \tau_i$ and every $\tau_j$ with a parent or a child of a stream of $\tau_i$}{\textsc{TryEnqueue}$(\tau_j)$}
  }
}
initialize $c_i = a_i = 0$, $\lambda_i = -\infty$, the heaps and the next jobs\;
call \textsc{TryEnqueue}$(\tau_i)$ for every task $\tau_i$\;
run \textsc{Worker} on every worker in parallel\;
\end{algorithm}

With the dependencies of Section~\ref{subsec:streams}, the simulator can run jobs out of order.  
The remaining issue is storage.
A stream can then run far ahead of its children, that is, many of its jobs finish before the jobs that consume their records start.
The simulator must store these records in the meantime, and their number is bounded only by the number of jobs of the stream, which grows with $H$.
To bound it, every stream keeps its records in a \emph{ring} with a fixed number of \emph{slots}, which its jobs reuse in turn.
A job that reuses a slot must then wait until the records in it have been consumed.
This adds a third kind of dependency.
If the dependencies had a cycle, the jobs on it would wait for each other, and none of them could start.
So we first define the rings and then choose their size such that the dependencies still form a directed acyclic graph.

Let the ring of a stream $\sigma$ have $K_\sigma$ slots.
Each slot holds the records of one job.
Job $J(\sigma, m)$ writes its records into slot $m \bmod K_\sigma$.
For $m \ge K_\sigma$, it replaces the records of $J(\sigma, m - K_\sigma)$.
These records are consumed by the jobs $J(\nu, m - K_\sigma)$ of the children $\nu$ of $\sigma$, which we call the \emph{slot dependencies} of $J(\sigma, m)$.
So $J(\sigma, m)$ can start only after its slot dependencies have finished.

The dependencies stay acyclic if every slot dependency of a job is released before the job.
Let $\sigma$ be a stream of a callback of $\tau_j$, and let $\delta_\sigma$ be the largest transport delay of the channels to its children, or $0$ if it has no children.
The records of $J(\sigma, m - K_\sigma)$ arrive at most $D_j + \delta_\sigma$ after its release, and they release the slot dependencies of $J(\sigma, m)$ when they arrive.
Job $J(\sigma, m)$ is released $K_\sigma T_\sigma$ after $J(\sigma, m - K_\sigma)$, so we need $K_\sigma T_\sigma > D_j + \delta_\sigma$ and choose
\begin{equation}
\label{eq:ring-size}
K_\sigma = \max\left(2,\ \left\lfloor \frac{D_j + \delta_\sigma}{T_\sigma} \right\rfloor + 1\right).
\end{equation}
The second term is the smallest such number, and the term $2$ lets a stream run at least one job ahead of its children.

\begin{lemma}
\label{lem:rings}
Every slot dependency of a job is released before the job.
\end{lemma}
\begin{proof}
The slot dependency $J(\nu, m - K_\sigma)$ of $J(\sigma, m)$, with $\nu$ on a channel $e$, is released at $r_\sigma(m) - K_\sigma T_\sigma + D_j + \delta_e$, which is before $r_\sigma(m)$ by~\eqref{eq:ring-size}.
\end{proof}

So every dependency of a job is released before it or comes before it in its task, and the dependencies still form a directed acyclic graph.

\begin{example}
\label{ex:rings}
In Example~\ref{ex:streams}, $K_{\sigma_1} = \max(2, \lfloor 10/10 \rfloor + 1) = 2$ and $K_{\sigma_2} = \max(2, \lfloor (25 + 4)/10 \rfloor + 1) = 3$.
So $J(\sigma_2, 3)$, released at $50$\,ms, reuses the slot of $J(\sigma_2, 0)$, and its slot dependency is $J(\sigma_3, 0)$, released at $49$\,ms.
Job $J(\sigma_1, 4)$, also released at $50$\,ms, waits for no job of $\tau_3$, directly or through other jobs, so it can run before $J(\sigma_4, 0)$, which is released at $20$\,ms.
\hfill$\diamond$
\end{example}

\subsection{The Simulation Algorithm}
\label{subsec:algorithm}

By Sections~\ref{subsec:streams} and~\ref{subsec:rings}, a job can start when its state, record and slot dependencies have finished.
Algorithm~\ref{alg:ooo} runs the jobs in this way on a pool of worker threads.
It has to find the jobs that can start and to make sure that each of them runs only once.
It does both without a list of jobs.

Every child stream has one parent, so the streams form trees whose roots are the root streams.
Before the run, the simulator searches these trees breadth first and keeps the streams with an offset before $H$.
It skips the children of all other streams, so the search always ends.
The simulator then computes the ranks and the ring sizes.
To find the next job of each task efficiently during the run, each task keeps the next job of each of its streams in a \textit{heap} ordered by key.
By Lemma~\ref{lem:streams}, the first job in the heap is the next job of the task.
The heap has at most one entry per stream, so no list of jobs is stored.

Each task $\tau_i$ also has three values.
Workers read and write them and the next job of the task atomically, in one total order.
The \emph{completion count} $c_i$ is the number of its finished jobs, and the \emph{completion key} $\lambda_i$ is the key of its last finished job.
The jobs of a task finish in the order of their keys, so a job of $\tau_i$ has finished if and only if its key is at most $\lambda_i$.
So one comparison of keys tells whether a dependency has finished.
To run every job only once, a worker must \emph{claim} a job before it runs it, and the \emph{claim count} $a_i$ is the number of claimed jobs of $\tau_i$.
\textsc{TryEnqueue}$(\tau_i)$ looks only at the next job $J(\sigma, m)$ of $\tau_i$, since the jobs of a task run in the order of its sequence.
If $a_i = c_i$, no job of $\tau_i$ is claimed and unfinished, so the state dependency of $J(\sigma, m)$ has finished (line~3).
Lines~5 and~6 check its record dependency $J(\pi, m)$ and its slot dependencies $J(\nu, m - K_\sigma)$.
If all have finished, the worker claims $J(\sigma, m)$ by a compare-and-swap (CAS) that changes $a_i$ from $c_i$ to $c_i + 1$ and fails if another worker has claimed a job of $\tau_i$ in the meantime (line~7).
It then pushes the job onto its local double-ended queue (deque).

Each worker takes the jobs from its local deque one at a time.
When its deque is empty, it steals a job from the deque of another worker, which balances the load among the workers (\textit{work stealing}~\cite{Blumofe1999}).
It runs a job $J(\sigma, m)$ of $\tau_i$ as in the semantics, with its release instant as the current time, and writes its records into slot $m \bmod K_\sigma$.
It then replaces $J(\sigma, m)$ in the heap by $J(\sigma, m + 1)$ if $m + 1 < M_\sigma$, sets the next job of $\tau_i$, sets $\lambda_i$ to $\kappa_\sigma(m)$ and increases $c_i$, in this order, so that a worker that reads the new $c_i$ also reads the new next job.
After that, the next job of $\tau_i$, a job that consumes a record of $J(\sigma, m)$, or a job that reuses a slot whose records $J(\sigma, m)$ consumed may be able to start.
So the worker calls \textsc{TryEnqueue} for $\tau_i$ and for every task that has a child or a parent of a stream of $\tau_i$.
Because of the total order, the worker that finishes the last dependency of a job sees that all of them have finished, so no job is missed.
The dependencies form a directed acyclic graph, so every job then runs once, and the simulation cannot deadlock.

\myparagraph{Complexity} Let $n_i$ be the number of streams of $\tau_i$, and let $d$ be the largest number of channels that enter or leave the callbacks of one task.
After a job of $\tau_i$ has run, the worker performs one heap operation in $O(\log n_i)$ time and calls \textsc{TryEnqueue} for at most $d + 1$ tasks.
Each call compares one pair of keys per dependency of the next job, in $O(1 + d)$ time.

\section{ROS~2 Integration}
\label{sec:layer}

We implemented the simulator of Section~\ref{sec:algorithm} as a lightweight layer for ROS~2 Jazzy.

\begin{figure}[t]
\centering
\definecolor{layerPale}{HTML}{FBE3DC}\relax
\begin{tikzpicture}[x=1pt, y=1pt, font=\fontsize{6.5}{7.5}\selectfont,
  bd/.style={draw=black!70, rounded corners=2pt, line width=0.4pt},
  cb/.style={draw=black!70, fill=white, rounded corners=1.5pt, line width=0.4pt,
    inner xsep=2pt, inner ysep=1.2pt, text height=4.5pt, text depth=1.4pt, minimum width=20pt},
  gh/.style={cb, draw=black!50, dashed, text=black!60},
  ar/.style={-{Latex[length=2.6pt,width=2.1pt]}, line width=0.45pt, draw=black!80},
  lb/.style={anchor=base west, inner sep=0pt},
  fl/.style={anchor=base west, inner sep=0pt, font=\fontsize{6}{7}\selectfont},
  ci/.style={circle, draw=black!70, line width=0.4pt, inner sep=0pt, minimum size=6.4pt},
  tp/.style={-{Latex[length=2.2pt,width=1.8pt]}, line width=0.4pt, draw=black!80}]
  \draw[bd, fill=white]     (0,41) rectangle (250,59);
  \draw[bd, fill=black!6]   (0,24) rectangle (250,38);
  \draw[bd, fill=layerPale] (0,0)  rectangle (197,14);
  \draw[bd, fill=black!6]   (201.5,0) rectangle (250,21);
  \node[lb] at (5,48.45) {Application};
  \node[lb] at (5,29.45) {\textit{rclcpp}};
  \node[lb] at (5,5.45)  {\textbf{Our simulator}};
  \node[lb, anchor=base] at (225.75,8.95) {\textit{rcl}, \textit{rmw}, DDS};
  \node[ci, fill=camera]   (a) at (82,54.3)  {};
  \node[ci, fill=camera]   (b) at (82,45.7)  {};
  \node[ci, fill=detector] (c) at (114,54.3) {};
  \node[ci, fill=detector] (e) at (146,50)   {};
  \node[ci, fill=planner]  (f) at (178,54.3) {};
  \node[ci, fill=planner]  (g) at (178,45.7) {};
  \node[ci, fill=planner]  (h) at (210,50)   {};
  \draw[tp] (a) -- (c);
  \draw[tp] (c) -- (e);
  \draw[tp] (b) -- (e);
  \draw[tp] (e) -- (f);
  \draw[tp] (e) -- (g);
  \draw[tp] (f) -- (h);
  \draw[tp] (g) -- (h);
  \node[cb] (ts) at (74,31)     {timers, subscriptions};
  \node[cb] (pb) at (128,31)    {publishers};
  \node[cb] (ck) at (174,31)    {ROS clock};
  \node[gh] (ex) at (225.75,31) {executors};
  \node[cb] (sim) at (74,7)  {workers};
  \node[cb] (rs)  at (128,7) {record slots};
  \node[cb] (ri)  at (174,7) {release instant};
  \draw[ar] (sim.north) -- (ts.south);
  \draw[ar] (pb.south)  -- (rs.north);
  \draw[ar] (ri.north)  -- (ck.south);
  \draw[ar] (rs.west)   -- (sim.east);
  \node[fl] at (76.5,17)  {callbacks};
  \node[fl] at (130.5,17) {messages};
  \node[fl] at (176.5,17) {time};
\end{tikzpicture}
\caption{Our simulator with ROS 2.}
\label{fig:layers}
\end{figure}
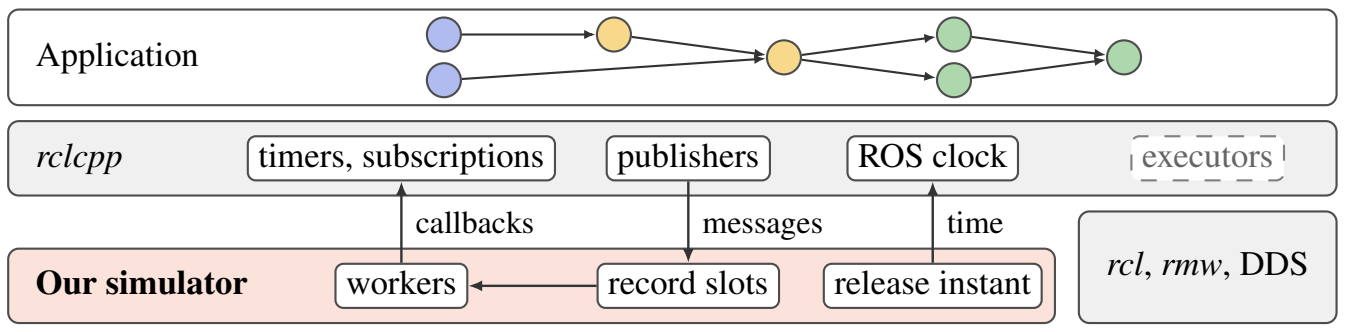

ROS~2 is organized in layers~\cite{Macenski2022}.
Fig.~\ref{fig:layers} shows them together with our simulator.
The nodes are written with a \emph{client library}, which is \textit{rclcpp} for C++~\cite{rclcppJazzy}.
It provides the timers, the subscriptions, the publishers and the ROS clock of a node, and its \emph{executors} decide when the callbacks run~\cite{ros2executors}.
Below it, the C library \textit{rcl} passes a published message through the interface \textit{rmw} to the \emph{middleware}, which delivers it to the subscriptions of the topic~\cite{rclJazzy}.
The middleware is usually an implementation of the Data Distribution Service (DDS)~\cite{OMGDDS}.

Our simulator is directly below \textit{rclcpp}.
To simulate the application as in the semantics of Section~\ref{subsec:semantics}, it runs the callbacks, carries the messages and gives the time itself.
To run a job, a worker of Algorithm~\ref{alg:ooo} calls its callback through the timer, or through the subscription once for every message of the consumed record, so no executor is started.
To take the published messages and to give the time, the simulator defines its own versions of the functions of \textit{rcl} that publish a message and that read the ROS clock.
They are part of the executable that loads the nodes, so \textit{rclcpp} calls them in place of the ones in \textit{rcl}, which stays unchanged.
A published message is then stored in serialized form in the slot of the running job, and the ROS clock returns the release instant of this job.
The simulator also turns off the direct delivery of messages inside one process, so that every message goes through a record.
The rest of ROS~2 is used as it is, without any change.

Using the simulator takes little effort.
The user does not write the application graph model of Section~\ref{subsec:graph}, because the simulator reads it from the nodes themselves.
It finds the timers and subscriptions of every node and connects publishers and subscriptions by topic name and message type.
It loads the nodes from their existing shared libraries, so they need no recompilation, or it creates them by the functions that construct them.
Besides the application, the simulator only needs the delays, which the user gives in files.

\section{Evaluation}
\label{sec:evaluation}

\begin{figure*}[t]
\centering
\includegraphics{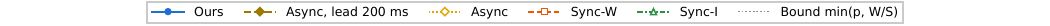}\\[2pt]
\begin{subfigure}[b]{0.245\textwidth}
\centering
\includegraphics{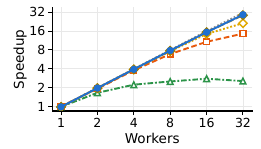}
\caption{Fleet-8 on 1 to 32 workers.}
\label{fig:eval-workers}
\end{subfigure}\hfill
\begin{subfigure}[b]{0.245\textwidth}
\centering
\includegraphics{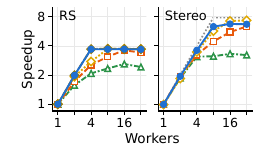}
\caption{RS and Stereo on 1 to 32 workers.}
\label{fig:eval-rs-stereo}
\end{subfigure}\hfill
\begin{subfigure}[b]{0.245\textwidth}
\centering
\includegraphics{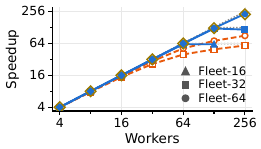}
\caption{Fleets on up to 256 workers.}
\label{fig:eval-node}
\end{subfigure}\hfill
\begin{subfigure}[b]{0.245\textwidth}
\centering
\includegraphics{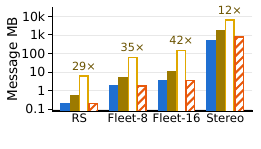}
\caption{Message bytes on 32 workers.}
\label{fig:eval-bytes}
\end{subfigure}
\caption{Speedup over sequential simulation and message bytes on the industrial use cases.}
\label{fig:eval-apps}
\end{figure*}

\begin{figure}[t]
\centering
\begin{subfigure}[b]{0.49\columnwidth}
\centering
\includegraphics{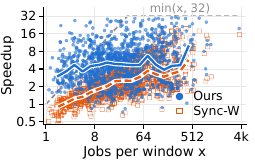}
\caption{Synthetic graphs.}
\label{fig:eval-synth}
\end{subfigure}\hfill
\begin{subfigure}[b]{0.49\columnwidth}
\centering
\includegraphics{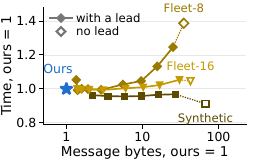}
\caption{Async with leads.}
\label{fig:eval-leads}
\end{subfigure}
\caption{(a) Speedup of the synthetic graphs, with medians as lines. (b) Time and message bytes of Async with leads, relative to ours. Both on 32 workers.}
\label{fig:eval-synth-leads}
\end{figure}

\subsection{Setup}

\myparagraph{Workloads and simulators}
We evaluate our simulator on three industrial use cases, the Autoware reference system (RS)~\cite{ReferenceSystem}, fleets of vehicles built from RS (Fleet-$N$), and a stereo vision pipeline (Stereo)~\cite{ImagePipeline,R2B2024}.
RS is a ROS~2 benchmark modeled on the autonomous driving software Autoware.
Fleet-$N$ runs $N$ copies of RS as vehicles that exchange detected objects, with different timer offsets.
Stereo runs the stereo vision nodes of image\_pipeline on recorded camera frames.
We also evaluate 1385 synthetic graphs, including randomly generated ROS~2 graphs and those derived from automotive benchmarks~\cite{Kramer2015,Hamann2017}.

We compare our simulator with sequential simulation, with the synchronous parallel simulations \texttt{Sync-I} and \texttt{Sync-W}, and with the asynchronous parallel simulation \texttt{Async}.
Sequential simulation runs one job at a time.
\texttt{Sync-I} and \texttt{Sync-W} run the jobs released in a time window in parallel and wait at a global barrier at its end~\cite{Fujimoto2000}.
\texttt{Sync-I} runs in parallel only the jobs released at the same instant, and waits at a barrier before the next instant.
\texttt{Sync-W} has windows as long as the smallest computation delay, so that the jobs of different tasks in a window do not depend on each other.
\texttt{Async} is a conservative simulation with null messages and without global barriers~\cite{ChandyMisra1979,Fujimoto1990}.
In \texttt{Async}, a task runs its next job as soon as it knows all records that the job reads, and an optional lead bounds how far a task may run ahead of its readers.

\myparagraph{Parameters and measurements}
We ran all simulators on ROS~2 Jazzy on a machine with two AMD EPYC 9755 processors and 256 cores, with each worker pinned to one core.
The computation delay of a node is its smallest timer period, or, if it has no timers, the smallest computation delay of the nodes that publish to it.

We measure how fast each simulator is and how many messages it holds.
The speedup of a simulator is the time of sequential simulation divided by its time.
To see how much of the parallelism of a workload a simulator uses, we compare its speedup with a bound that depends only on the workload.
Let $W$ be the total computation time of all jobs and $S$ that of the longest chain of jobs in which each job needs the state or a record of the job before it, both measured in sequential simulation.
The jobs of such a chain run one after another, so any simulator takes at least $S$, and $p$ workers take at least $W/p$.
The speedup on $p$ workers is therefore at most $\min(p, W/S)$, which we call the bound.
The ratio $W/S$ is the average number of jobs that the workload can run at the same time.
This ratio is 3.7 for RS and 7.9 for Stereo, and it grows in proportion to the number of vehicles in the fleets, up to 252 for Fleet-64.
The message bytes of a simulator are the most bytes of messages that it holds at one time.

\subsection{Results}

\myparagraph{Speedup}
On Fleet-8, our simulator follows the bound and is 29.1 times faster than sequential simulation on 32 workers, which is 93\,\% of the bound (Fig.~\ref{fig:eval-workers}).
\texttt{Sync-W} reaches 14.6, and \texttt{Sync-I} stays below 3, because the vehicles have different offsets and few jobs share an instant.
When all vehicles have offset 0, \texttt{Sync-I} and \texttt{Sync-W} reach speedups of 15.1 and 18.0, and our simulator reaches 28.9.
Aligning the release instants therefore improves the speedup of synchronous simulation, while the speedup of our simulator stays almost the same.
With transport delays of 0 to 100\,ms between the vehicles of Fleet-8, our simulator stays between 29.0 and 30.9 times faster than sequential simulation on 32 workers.
RS and Stereo have bounds of only 3.7 and 7.9 (Fig.~\ref{fig:eval-rs-stereo}).
Our simulator reaches the bound of RS on 4 workers, where \texttt{Async} and \texttt{Sync-W} reach 2.8 and 2.5.
On Stereo, it reaches 6.7, and \texttt{Async} reaches 7.4.

\myparagraph{Scaling}
On the whole machine, Fleet-64, with 1600 nodes and 566\,353 jobs in 10\,s, takes 228\,s in sequential simulation (Fig.~\ref{fig:eval-node}).
Our simulator takes 1.0\,s on 256 workers, so it is 228 times faster, which is 91\,\% of the bound of 252.
\texttt{Sync-W} reaches 89, so our simulator is 2.6 times faster.
In every window, \texttt{Sync-W} waits for the task with the most work, and our simulator does not.

\myparagraph{Message bytes}
\texttt{Sync-W} holds as many message bytes as our simulator, or 1.5 times as many on Stereo, but \texttt{Async} holds 12 to 42 times as many, since a task of \texttt{Async} can run far ahead of its readers and its records then wait in queues (Fig.~\ref{fig:eval-bytes}).
On Stereo, \texttt{Async} is about as fast, but it holds up to 6.3\,GB of messages at one time, while our simulator holds at most 0.5\,GB.
A lead bounds the queues, but it has to be chosen for each workload (Fig.~\ref{fig:eval-leads}).
On Fleet-8, \texttt{Async} without a lead is 1.4 times slower and holds 35 times as many message bytes, while leads from 25 to 200\,ms make it as fast as our simulator with 1.4 to 2.9 times its message bytes.

\myparagraph{Synthetic graphs}
Fig.~\ref{fig:eval-synth} shows the speedups of our simulator and \texttt{Sync-W} on the synthetic graphs with 32 workers, plotted against the average number $x$ of jobs in a nonempty window of \texttt{Sync-W}.
If all jobs take the same time, the speedup of \texttt{Sync-W} is at most $\min(x,32)$.
Across all 1385 synthetic graphs, our simulator is 1.98 times as fast as \texttt{Sync-W} on average (geometric mean).
\texttt{Async} is about as fast as our simulator but holds 45 times as many message bytes.
With leads of 0.25 to 8 times the longest timer period, \texttt{Async} still holds 2.2 to 27 times as many message bytes (Fig.~\ref{fig:eval-leads}).

\section{Related Work}
\label{sec:related}

Safety-critical embedded systems need behavior that is deterministic and predictable~\cite{Lee2021}.
The time-triggered architecture was designed for this, with computation and communication at predefined time instants~\cite{Kopetz2003}.
Henzinger et~al.\ developed this idea into the logical execution time (LET)~\cite{henzinger2003giotto}, which our coordination scheme follows.
AUTOSAR supports LET~\cite{AUTOSARTimex}.
LET fixes the instants of reads and writes on one machine, but messages between machines take a varying time.
System-level LET~\cite{Gemlau2021} therefore gives the communication between machines a fixed delay, as our transport delays do.
The Adaptive Platform of AUTOSAR~\cite{AUTOSARAP} comes closest to ROS~2, since its applications also run on POSIX and communicate by publish and subscribe.
Menard et~al.\ show that it is not deterministic by itself and add a coordination layer to it~\cite{Menard2020}.
Bellassai et~al.\ instead bring system-level LET to it~\cite{Bellassai2025} and to its publish-subscribe communication~\cite{Bellassai2025APLET}.
In both works, LET applies only to time-triggered tasks.
Our scheme also applies to event-driven callbacks.
An empty record tells such a callback that no message comes.
In this way, the absence of a message also carries information, as Lamport observed for bounded message delays and synchronized clocks~\cite{Lamport1984}.

Three recent approaches make the simulation or the execution of ROS~2 applications deterministic.
Sagmeister et~al.\ run the callbacks of all nodes sequentially on a single thread with a simulated clock~\cite{Sagmeister2026}.
Otto uses an orchestrator that forwards the messages between nodes in an order that it builds during the run~\cite{Otto2023}.
Teper et~al.\ map a ROS~2 application to Lingua Franca~\cite{Teper2026}, a coordination language based on the deterministic reactor model~\cite{Lohstroh2021}.
Each node becomes a reactor, a component that reacts to timestamped events, and each callback becomes one of its reactions.
The runtime of Lingua Franca then calls the unmodified callbacks in timestamp order.
Kwok et~al.\ instead build a robotic middleware on Lingua Franca to replace ROS~2, and show how to move ROS~2 programs to it~\cite{Kwok2025}.
Although Lingua Franca can also express LET~\cite{LeeLohstroh2022}, its runtime is synchronous by default and completes all reactions at one timestamp before it advances to the next.
By default, none of the three runs callbacks of different timestamps in parallel, since a callback may or may not publish and which callback runs next is known only during the run.
With empty records, the jobs and their dependencies are known before the run, so our simulator runs jobs out of order and in parallel.
An empty record plays the role of a null message in conservative parallel simulation~\cite{ChandyMisra1979,Fujimoto1990}, but it is part of our coordination scheme and not an extra message of the simulator.

Henzinger and Kirsch describe semantics-preserving compilation of LET programs for a single processor~\cite{Henzinger2002EmbeddedMachine}.
Biondi and Di Natale implement LET on multicore platforms~\cite{Biondi2018}.
We use LET to coordinate ROS~2 callbacks, defining the application behavior that our simulator reproduces.
The final step would be to generate code that preserves validated behavior on the target platform.

\section{Conclusion}
\label{sec:conclusion}

We have shown that fixing when every callback reads and writes makes a ROS~2 application deterministic.
With empty records, it also fixes the jobs and their dependencies before the run.
Based on this, our simulator runs the jobs out of release order and in parallel.
It reached 91\,\% of the theoretical speedup bound on 256 cores.

Our scheme assumes a fixed application structure and at most one publisher per topic within each node.
As in LET, deferring outputs until their computation delays elapse adds end-to-end latency.
Future work includes relaxing these restrictions and choosing delays to meet application timing requirements.

\clearpage
\bibliographystyle{IEEEtran}
\bibliography{ref}

\end{document}